\documentclass[conference]{IEEEtran}
\IEEEoverridecommandlockouts
\usepackage{cite}
\usepackage{amsmath,amssymb,amsfonts}
\usepackage{algorithmic}
\usepackage{graphicx}
\graphicspath{ {images/} }
\usepackage{subfigure}
\usepackage{textcomp}
\usepackage{xcolor}
\usepackage[ruled,linesnumbered]{algorithm2e}
\def\BibTeX{{\rm B\kern-.05em{\sc i\kern-.025em b}\kern-.08em
    T\kern-.1667em\lower.7ex\hbox{E}\kern-.125emX}}

\begin{document}

\title{Zero-Shot Learning Based on Knowledge Sharing\\}
\author{\IEEEauthorblockN{Zeng Ting, Xiang Hongxin, Xie Cheng\IEEEauthorrefmark{1}, Yang Yun, Liu Qing } \IEEEauthorblockA{School of Software, Yunnan University, China\\
{\footnotesize \textsuperscript{*}The Corresponding Author xiecheng@ynu.edu.cn}
}}

\maketitle

\begin{abstract}
Zero-Shot Learning (ZSL) is an emerging research that aims to solve the classification problems with very few training data. The present works on ZSL mainly focus on the mapping of learning semantic space to visual space. It encounters many challenges that obstruct the progress of ZSL research. First, the representation of the semantic feature is inadequate to represent all features of the categories. Second, the domain drift problem still exists during the transfer from semantic space to visual space. In this paper, we introduce knowledge sharing (KS) to enrich the representation of semantic features.  Based on KS, we apply a generative adversarial network to generate pseudo visual features from semantic features that are very close  to the real visual features. Abundant experimental results from two benchmark datasets of ZSL show that the proposed approach has a consistent improvement.
\end{abstract}

\begin{IEEEkeywords}
Zero-Shot learning, semantic representation, semantic augmentation, domain drift
\end{IEEEkeywords}

\section{Introduction}
The ways to obtain training data are more comfortable since there are more and more annotated datasets available online in recent years. However, how to extend the classification to a large number of classes if the training samples of some classes do not available is still a critical challenge. The interesting thing is, compared to the computer,  the human can always infer the unseen objects even he/she did not ever see the objects before. Indeed, human has a unique ability to use the existing knowledge to "imagine" the unseen object. Inspired by the idea, Zero-Shot learning\cite{5206594,Xian2017Zero,palatucci2009zero} has become a common method to classify unseen objects without seeing the samples.

The main challenge of ZSL is how to recognize novel classes without accessing any labeled sample of these classes. So far, many methods have emerged to deal with unseen classes without samples. Some methods \cite{5206594,6974493,10.1007/978-3-642-15555-0_10} use the text attribute to predict  unseen objects but do not get a good result. Some other methods consider the ZSL task as a mapping problem\cite{Kodirov_2017_CVPR}. It maps the relationships between semantic space and visual space. However, these mapping-based approaches would encounter the domain drift problem that has many limitations. The recent work\cite{Zhu_2018_CVPR} uses the generative adversarial network (GAN) to generate unseen visual features. It achieves the highest accuracy in two benchmarks of ZSL. However, the approach suffers from both  inadequate semantic representation and domain drift problem.

In this paper, we focus on two critical problems in ZSL. The first one is domain shift\cite{fu2015transductive}. For example, Zebra and Pig share the same 'hasTail' semantic attribute, but the visual appearance of their tails is different. Similarly, Pig's other attributes features are visually different from the corresponding attributes features of Zebra. Therefore, such a difference will reduce the recognition rate of ZSL. The second problem is inadequate semantics. For models, it is challenging to recognize unseen objects only with semantic knowledge learned from seen classes. To address these issues, the KS method is used to augment semantic representation. By merging several class texts with high similarity, we can introduce the text description of unseen classes into seen classes, to reduce the knowledge gap between seen classes and unseen classes. Our method for zero-shot learning that achieved 43.95\% and 37.02\% on SCS-split and 11.31\% and 9.26\% on SCE-split on Caltech UCSD Birds-2011(CUB)\cite{WahCUB_200_2011} and North America Birds(NAB)\cite{Horn2015Building} datasets separately.
\\

\section{Related Work}

\subsection{Zero-Shot Learning}
The classic ZSL methods start with training seen classes and gain knowledge about attributes, then inference unseen classes and classify unseen objects via known knowledge. In 2009, ZSL's pioneering work \cite{5206594} proposed Direct Attribute Prediction(DAP). It learns the attribute classifier firstly and then seeks the most promising unseen class. There are some similar works\cite{10.1007/978-3-642-15555-0_10, NIPS2014_5290,6974493}. These works either ignore the relation between attributes or map the visual feature space directly to the semantic space, reducing the accuracy of ZSL greatly. In 2018, \cite{Zhu_2018_CVPR} combined GAN and ZSL and transformed ZSL into an imagination problem. It composes visual features rather than synthetic images through a conditional GAN while ensuring the unseen distinguish between classes and the diversity of synthetic data with the class. Given this, we combined GAN and KS, which greatly improved performance.

\subsection{Generative Adversarial Network} 
GAN \cite{NIPS2014_5423} mainly consists of two models: 1) a generative model G, and 2) a discriminative model D that contest against each other. Since GAN was born, it has achieved excellent performance on ZSL because of its ability to generate realistic images\cite{NIPS2014_5423} \cite{SEPARATELY20173C}. However, the instability in training and mode collapse become its main shortcomings. To alleviate these problems and improve the quality of artificial samples, many methods have been proposed. M. Arjovsky proposed WGAN\cite{Arjovsky2017Wasserstein} and WGAN-GP \cite{Gulrajani2017Improved} to optimize GAN on an approximate Wasserstein distance by enforcing 1-Lipschitz smoothness. LS-GAN \cite{Mao2016Least} offers a simple but effective solution by replacing the cross-entropy loss of GAN with a least-square loss that pushes the scores of real and fake samples to different decision boundaries. David Berthelot et al. proposed BEGAN\cite{berthelot2017began} in 2017, which can train with a simple network and solve model collapse. Among these excellent GANs, we choose the most popular GAN variant, ACGAN\cite{OdenaConditional}, as the basic architecture of our model in this paper. ACGAN will be detailed in Chapter \ref{approach}. ACGAN has gained great achievement in training stability and generating a result.

\section{The Approach}
\label{approach}
\begin{figure*}
\centering
\includegraphics[width=16cm,height=8cm]{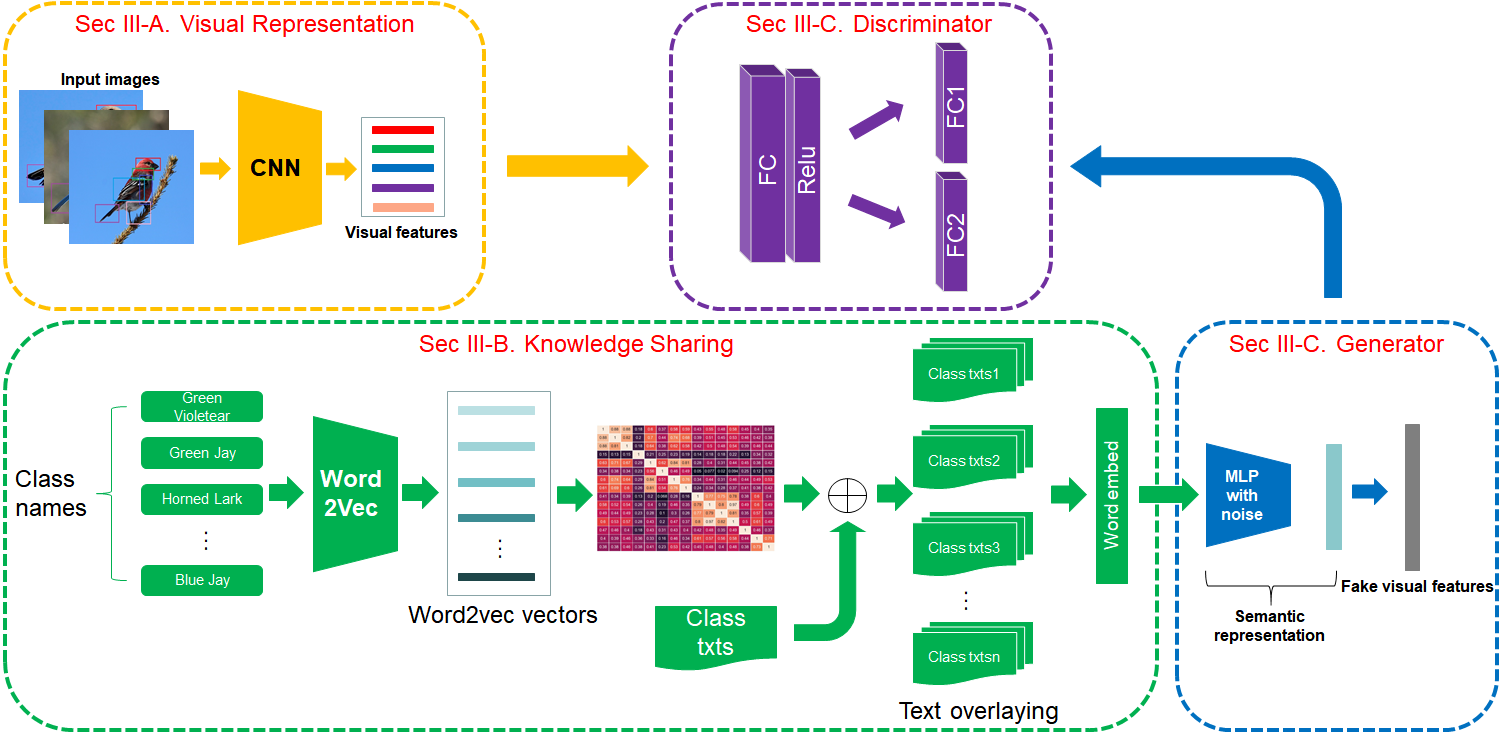}
\caption {Overall architecture: Fast RCNN with the vgg16 backbone is used to detect every part of the input image (head, tail, foot,...). Each detected part is extracted by the CNN model. The extracted features are spliced together to form real visual features (Sec \ref{Visual Feature Extraction}). Word2vec is used to map class names to 100-dimensional word vectors. The class similarity matrix consists of the Euclidean distance of the word vector corresponding to a pair of class names. Then, the text of each class is updated to the total text of the top k classes. These new texts are embedded into the semantic vector by TF-IDF method (Sec \ref{Semantic Feature Extraction}). The generator is used to map semantic vectors to pseudo visual features (SEC \ref{Generator and Discriminator}) with the same dimension as the real visual features. Its purpose is to use pseudo visual features to cheat the discriminator. Finally, the discriminator is trained to use real visual features and pseudo visual features (SEC \ref{Generator and Discriminator}). The purpose of the discriminator is to correctly classify the visual features and distinguish the true and false world features.}
\label{modelFrame}
\end{figure*}

As described in \cite{Xian2017Zero,Ba2015Predicting,Pambala2019Unified}, most types of ZSL methods can be unified into an embedding-based framework. We wanted to find a semantic-to-visual mapping that would give the model an ability to ``imagine". Similar works\cite{zhang2017learning,shigeto2015ridge,yang2014unified,lei2015predicting,Akata_2015_CVPR,akata2016multi,romera2015embarrassingly,Zhu_2018_CVPR} show that this approach yielded optimistic results. Therefore, the core of the proposed method is to design a scheme that can transform semantic features into visual features better and improve the recognition accuracy of unseen classes. Our method, like \cite{Zhu_2018_CVPR}, adopts the framework of ACGAN to transform semantic features into visual features. Fig.\ref{modelFrame} shows the overall architecture. Section \ref{Visual Feature Extraction} introduces the process of extracting visual features from input images. Section \ref{Semantic Feature Extraction}  introduces the process of extracting semantic features from texts. Section \ref{Generator and Discriminator} details the structure of GAN. Finally, Section \ref{Training and Testing} describes the steps to train model and test.


\subsection{Visual Representation}
\label{Visual Feature Extraction}
Firstly, regions of input images were detected through fast-RCNN with VGG16 backbone. Then, the detected regions are input into the visual encoder subnet, and it will eventually be encoded into 512-dimensional feature vectors for each part. The visual features are extracted by concatenating the 512 dimensional features of each part. 

\subsection{Semantic Representation}
\label{Semantic Feature Extraction}

\begin{figure}
\
\includegraphics[width=8cm,height=4cm]{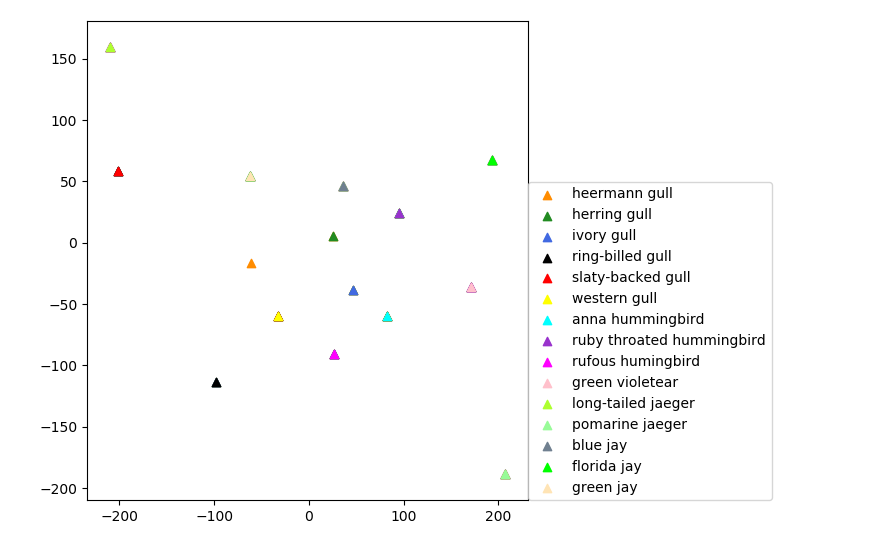}
\caption {t-SNE visualization of class features, which is embedded by the word2vec method, selected from classes in CUB randomly. The closer the two categories are, the higher their similarity are.}
\label{word2vecSimWithGreenJay}
\end{figure}

\subsubsection{Preprocessing} Some preprocessing works, which are removing stop words and porter stemmor \cite{Porter2013An}, need to be carried out before Wikipedia text is transformed into semantic features.

\subsubsection{Knowledge Sharing} Fig.\ref{word2vecSimWithGreenJay} shows a scatter diagram of class names encoded by word2vec mapped to two-dimensional space by the t-SNE method. It can be seen that green jay and blue jay are relatively close to each other because they have the same parent, jay. Furthermore, they also have certain similarities in environment and habits. While green jay and pomarine jeager are very far apart due to different parents. However, not belonging to the same parent class, the more similar. For example, green jay and green violetear do not belong to the same parent class and are closer to each other than florid jay because of similar habits and same colors. Therefore, super parent, color and habit all can decide the distance of classes. Next, we will describe the process in detail. For each class $c \in C$, we first use word2vec to embed their class names into a vector space and get $embed(c)$. According to the embedding vector, a class similarity matrix $SM \in R^{n \times n}$ can be obtained by calculating the similarity of each class. For each class $c$, we rank the similarity of each class and combine the text of top k classes. Finally, the TF-IDF method is used to encode each class of text as the input of the method to extract semantic features. Pseudo-code for the KS method is provided in algorithm \ref{KSAlgorithm}.

\begin{algorithm}
    \caption{KS algorithm}
    \label{KSAlgorithm}
    \KwIn{Total number of unique labels $n$, Similarity descending ranking $k$, text encoder $\Phi$,word2vec model $w2v$, class name of each class $E$, original wiki articles of each class $A_o$ \;}
    \KwOut{similarities matrix between classes $SM \in R^{n \times n}$, similary-based wiki articles $A_s$, encoded text vectors $\Phi(A_s)$ \;}
    Initialize $M_s=\emptyset, A_s=A_o $\;

      \For{$i=1;i \le n;i++$}
      {
      $s1=w2v(E_i)$\;
      \For{$j=1;j\le n;j++$}
        {
        $s2=w2v(E_j)$\;
        $similarity = distance(s1, s2)$\;
         $M_s[i][j]=similarity$\;
       }
       Rank $M_s[i]$ in descending order and select the top $k$ classes, and mark them as $k_c$\;.
      \For{$c=1;c \le k;c++$}
      {
      Add the wiki article of class $k_c$ to $A_s^i$\;
      }
      }
     $\Phi(A_s) \leftarrow$ encoding $A_s$ with TFIDF encoder\;
    return $M_s$, $A_s$, $\Phi(A_s)$ \;
\end{algorithm}

\subsection{Generator and Discriminator}
\label{Generator and Discriminator}

The generator ($G_\theta$ for short, $\theta$ is parameters of the generator) is a multi-layer perceptron consisting of a full connection layer with random noise $z$, a full connection layer with LeakRelu activation function and a full connection layer with Tanh activation function. The semantic features $T$, obtained by describing the method in Section \ref{Semantic Feature Extraction}, are used as input to the generator. Fake visual features  $\widetilde{x} $ can be generated by $ G_{ \theta } (T,z)$. 

The discriminator ($D_\omega$ for short) accepts the fake visual features from $G_\theta$ and the real visual features extracted from the real image as input. These features are input to a fully connected layer with ReLu activator for further feature extraction. Finally, the two subnetworks are used to judge whether the visual features are real or fake, and the corresponding classification labels of the visual features.



\subsection{Training and Testing}
\label{Training and Testing}
The complete steps to train and test the model are divided into four steps, followed by a detailed description of each step:

\textbf{Step1. Training Word2vec Model:} Before implementing the KS algorithm, out of vocabulary (OOV) exists in the pre-trained word2vec model, the word2vec model requires retraining using Wikipedia text from English wiki pages, CUB and NAB.

\textbf{Step2. Constructing Similarity-Based Text:} Using the KS method to construct new similarity-based text. See the algorithm \ref{KSAlgorithm} for details.

\textbf{Step3. Training GAN Model:} To train GAN model, $x_i$ and $\Phi (A_s^ i)$ forms visual-semantic pairs as input of GAN. The discriminator is used to distinguish the real and fake input samples and to predict the labels of visual features. The generator is optimized to fool the discriminator.

\textbf{Step4. Testing for Zero-Shot:} After model training, it is easy to generate pseudo-visual features of $G_\theta (T, z)$ according to the description text of a certain category $c$. Text features $t_u$ and random noises $z$ are used to generate $n$ pseudo-visual features for each category as training samples. These training samples are used to train a machine learning model. For visual features of each unseen class image, the classifier is used to classify it directly.

\section{Experiments}
\subsection{Datasets Settings}
\begin{figure}
\
\includegraphics[width=8cm,height=8cm]{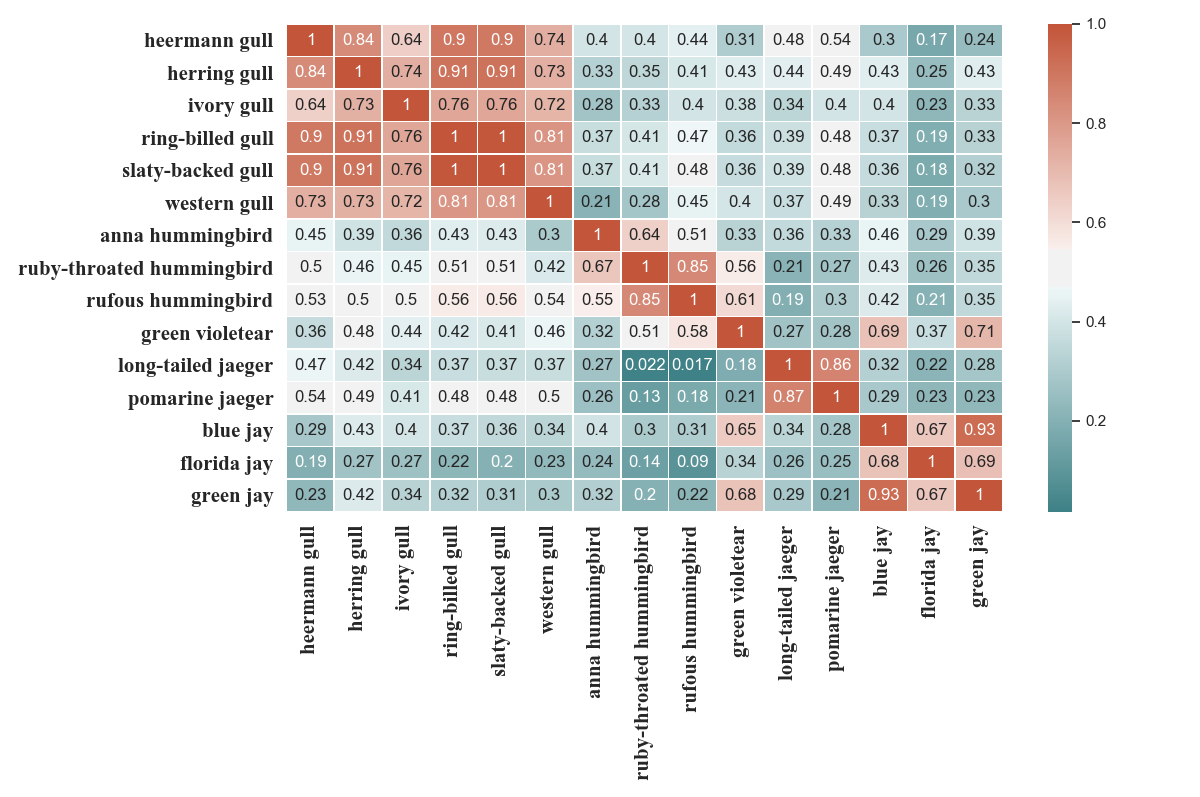}
\caption {Similarity matrix for class names. Each value in the figure represents the similarity between the category name corresponding to the X-axis and the Y-axis. These similarity values are calculated using word2vec and Euclidean distance.}
\label{similarity matrix}
\end{figure}

Two benchmark  datasets, Caltech UCSD Birds-2011(CUB) and North America Birds(NAB), are used to make a comparison between our approach and the state-of-the-art approaches. There are 11788 images from 200 classes in CUB dataset. While, the NAB dataset contains more images, with 49,562 images in 1011 classes. Besides, the original texts from English Wikipedia-v01.02.2016 are used in this paper. Word2vec method is used to embed class names into semantic vectors, and Euclidean distance is used to calculate the similarity between semantic vectors. The similarity matrix of CUB data classes is shown in Fig.\ref{similarity matrix}. Categories with the same superclass have high similarity. For example, the similarity between heermann gull and herring gull is 84\% since they have the same superclass, gull. However, heermann gull and florida jay, do not belong to a superclass, are only 17\% similar. It is worth noting that the similarity of different superclasses is not necessarily lower than that of the same superclass. The similarity of green Jay and Florida Jay (67\%) is lower than that of green Jay and green violetear (68\%).

Super-Category-Shared splitting (SCS) and Super-Category-Exclusive splitting (SCE) are two commonly used split settings \cite{Elhoseiny2017Link,huzero,ji2018stacked}. In the SCS-split, there must be a seen class that belongs to the same superclass as the unseen class. For example, ``Laysan Albatross" and ``Black-Footed Albatross" are in the seen classes and the unseen classes respectively, they have the same superclass, ``Albatross". In the SCE-split, there is no shared superclass between seen classes and unseen classes. In the NAB dataset, ``Scarlet Tanager" is selected as the seen classes, and all classes with ``Tanager" as superclass will be selected as seen classes. 

\subsection{Training Details and Parameters}
\subsubsection{Textual Features}
Firstly, the KS method is used to superimpose Wikipedia text of each class. Then, Wikipedia text processed by the KS method needs to be cleaned up by removing stop words, porter stemmer, and tokenized methods. Finally, Wikipedia text is represented using Term Frequency-Inverse Document Frequency(TF-IDF). The extracted TF-IDF feature dimensions are 7551 and 13217 for the CUB2011 and NAB Wikipedia data sets, respectively. \\
\subsubsection{Visual Features}There are seven parts: (1) head, (2) back, (3) belly, (4) breast, (5) leg, (6) wing and (7) tail in each input image. These parts are divided for capturing the different characteristics of birds. For each part of birds, a 512-dimensional vector can be attained by using MLPs with two hidden layers(each with size 512). For the CUB2011 dataset, seven parts of birds are all used as visual features, while in NAB, the ``leg" part is deleted. Therefore, the feature dimensions extracted from CUB2011 and NAB datasets are 3584 and 3072, respectively.\\

\subsubsection{Model Setting} 
Different super parameters k (k = 1,2,3,4,5) are set for the KS method in Section \ref{superParamStudy}. We use Adam as the optimizer of the model, with its default parameters $\beta_1=0.9, \beta_2=0.999$ and learning rate $\eta=0.001$. We train our model until the maximum number of iterations $maxiter=10000$ and the size of each batch of data $batchsize=1000$ is set. Every 40 iterations, KNN (k = 20) is trained to evaluate the performance of the method in seen and unseen classes. The highest generalized accuracy of seen class corresponding to the unseen class accuracy is selected as the final result. We usually use 0.5 as the threshold of our classification results. If the classification probability is higher than 0.5, it is true, and if it is less than 0.5, it is false. The generalization accuracy uses different thresholds as probability results to calculate the average accuracy under different thresholds.
\subsection{Evaluation Metric}
The popular Top-1 accuracy is used to evaluate the predictive performance of our model in ZSL. Top-1 accuracy is a widely used evaluation metric\cite{Meng2018Self,Xian2017Zero}. It calculates the proportion of correctly labeled samples to the total test samples and chooses the highest prediction probability as  the final result.

\subsection{Hyper Parameter Study}
\label{superParamStudy}

\begin{figure}
\subfigure[CUB Dataset]{
\begin{minipage}[t]{0.5\linewidth}
\centering
\includegraphics[height=3cm,width=4cm]{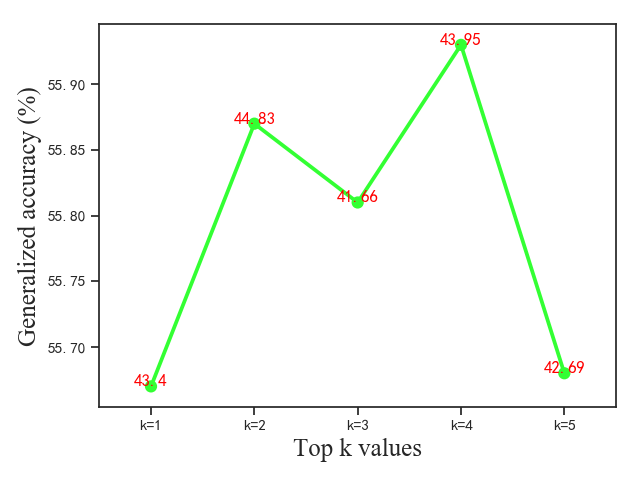}
\end{minipage}%
}%
\subfigure[NAB Dataset]{
\begin{minipage}[t]{0.5\linewidth}
\centering
\includegraphics[height=3cm,width=4cm]{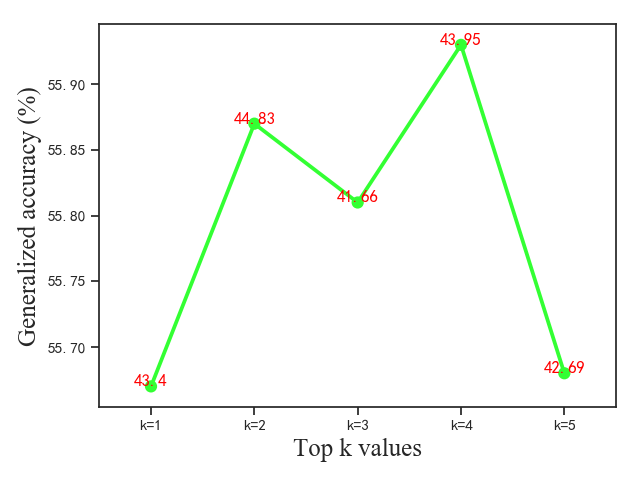}
\end{minipage}%
}%
\caption{Generalized Accuracy on two benchmarks datasets with SCS-split}
\label{superParamStudySCS}
\end{figure}

\begin{figure}
\subfigure[CUB Dataset]{
\begin{minipage}[t]{0.5\linewidth}
\centering
\includegraphics[height=3cm,width=4cm]{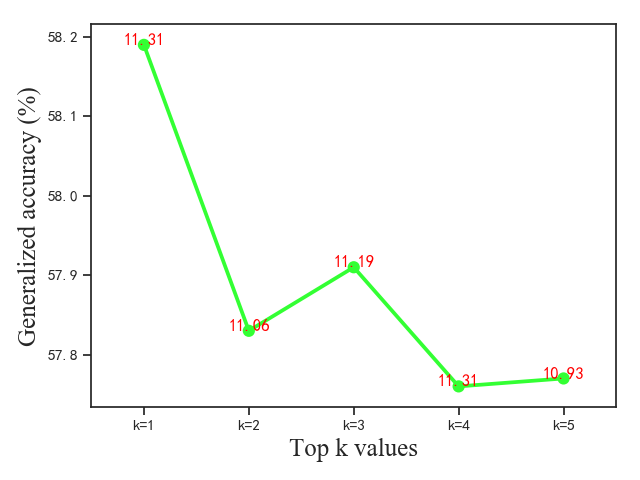}
\end{minipage}%
}%
\subfigure[NAB Dataset]{
\begin{minipage}[t]{0.5\linewidth}
\centering
\includegraphics[height=3cm,width=4cm]{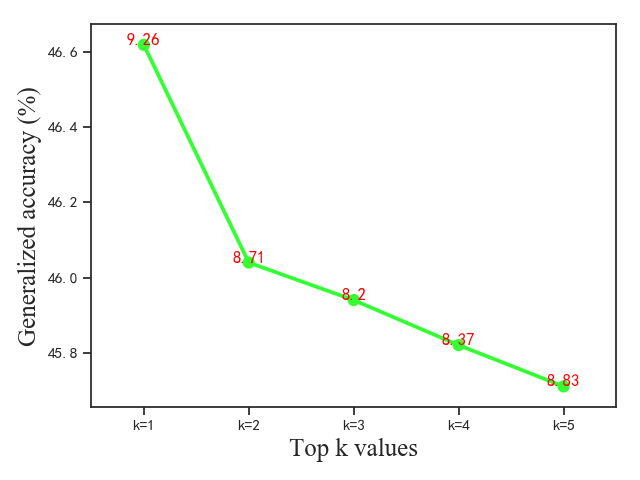}
\end{minipage}%
}%
\caption{Generalized Accuracy on two benchmarks datasets with SCE-split}
\label{superParamStudySCE}
\end{figure}

Different hyperparameters are set to evaluate their effect on the model. Fig.\ref{superParamStudySCS}, \ref{superParamStudySCE} show the generalized accuracy curves on the two benchmark datasets with varying methods of splitting and different hyperparameters. 
As shown in Fig.\ref{superParamStudySCS} and Fig.\ref{superParamStudySCE}, the value of hyperparameter setting is represented with the horizontal axis, the generalization accuracy of the seen classes is represented with the vertical axis, and the corresponding accuracy of the seen class is represented with red value in the curve. The hyperparameters with the highest generalization accuracy will be selected as the parameter of the model. Therefore, the model needs different hyperparameters values for different settings. For CUB, the $k$ value of $top-k$ is set as 4 and 1 in SCS and SCE cases, respectively. While for NAB, the $k$ value of $top-k$ is 3 and 1 in the case of SCS and SCE, respectively. $top-k$ represents the sharing of k classes of text. 
\subsection{Comparative Methods}
\par Ten newest methods are used to compare with KS for comparing: MCZSL\cite{akata2016multi}, WAC-Linear\cite{Elhoseiny2014Write}, WAC-Kernel \cite{Elhoseiny2016Write}, ESZSL\cite{romera2015embarrassingly},  SJE \cite{Akata_2015_CVPR}, ZSLNS\cite{Qiao2016Less}, SynCfast\cite{Changpinyo2016Synthesized}, SynCOVO \cite{Changpinyo2016Synthesized}, ZSLPP \cite{Elhoseiny2017Link}, and GAN-ZSL\cite{Zhu_2018_CVPR}. All the comparison methods used the same splits. Therefore, it is convenient to cite the results from\cite{Zhu_2018_CVPR} and the literature for fair comparisons. The method adopts two datasets segmentation methods of SCE and SCS. It performs performance evaluation on two benchmark datasets. As shown in Table \ref{top1AccWithOtherModels},  it is obvious that our method has achieved the most advanced result. It is worth noting that ZSL still has some challenges on SCE. Nevertheless, our method still improved by 9.8\% and 7.7\% than GAN-ZSL on the CUB dataset and NAB dataset with SCE-split, respectively. Similarly, our approach has yielded considerable results on SCS.

\newcommand{\tabincell}[2]{\begin{tabular}{@{}#1@{}}#2\end{tabular}}
\begin{table}[!htb]
\center
\caption{Top-1 accuracy (\%) on CUB and NAB datasets with two split settings}
\begin{tabular}{|l|l|l|l|l|}
\hline
                   & \multicolumn{2}{l|}{\textbf{CUB}} & \multicolumn{2}{l|}{\textbf{NAB}} \\ \hline
\textbf{Methods}            & \textbf{SCS}         & \textbf{SCE}        & \textbf{SCS}         & \textbf{SCE}        \\ \hline
MCZSL \cite{akata2016multi}      & 34.7        & -          & -           & -          \\ \hline
WAC-Linear \cite{Elhoseiny2014Write} & 27.0        & 5.0        & -           & -          \\ \hline
WAC-Kernel \cite{Elhoseiny2016Write} & 33.5        & 7.7        & 11.4        & 6.0        \\ \hline
ESZSL \cite{romera2015embarrassingly}     & 28.5        & 7.4        & 24.3        & 6.3        \\ \hline
SJE \cite{Akata_2015_CVPR}        & 29.9        & -          & -           & -          \\ \hline
ZSLNS \cite{Qiao2016Less}     & 29.1        & 7.3        & 24.5        & 6.8        \\ \hline
SynCfast \cite{Changpinyo2016Synthesized}   & 28.0        & 8.6        & 18.4        & 3.8        \\ \hline
SynCOVO \cite{Changpinyo2016Synthesized}    & 12.5        & 5.9        & -           & -          \\ \hline
ZSLPP \cite{Elhoseiny2017Link}     & 37.2        & 9.7        & 30.3        & 8.1        \\ \hline
GAN-ZSL \cite{Zhu_2018_CVPR}    & 43.7        & 10.3       & 35.6        & 8.6        \\ \hline

Ours               & \textbf{43.95}       & \textbf{11.31}      & \textbf{37.02}            & \textbf{9.26}           \\ \hline
\end{tabular}
\label{top1AccWithOtherModels}
\end{table}

\subsection{Generalized Zero-shot Learning}
\begin{figure}
\subfigure[CUB Dataset]{
\begin{minipage}[t]{0.5\linewidth}
\centering
\includegraphics[height=3cm,width=4cm]{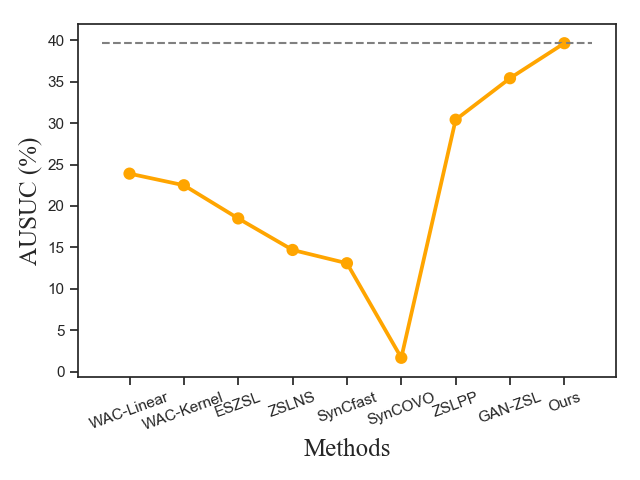}
\end{minipage}%
}%
\subfigure[NAB Dataset]{
\begin{minipage}[t]{0.5\linewidth}
\centering
\includegraphics[height=3cm,width=4cm]{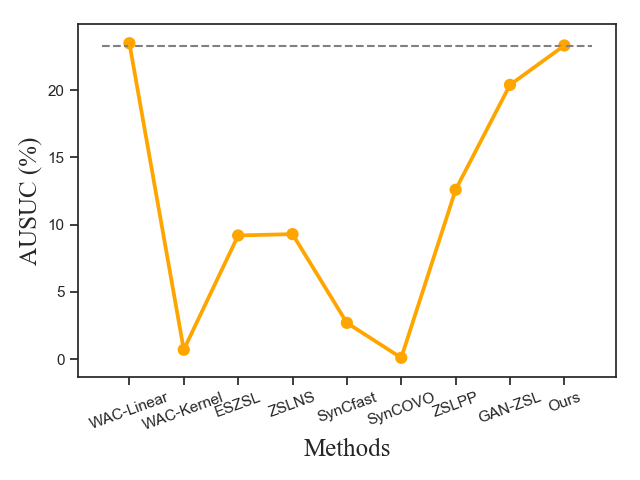}
\end{minipage}%
}%
\caption{AUSUC on two benchmarks datasets with SCS-split}
\label{SCS_AUSUC}
\end{figure}

\begin{figure}
\subfigure[CUB Dataset]{
\begin{minipage}[t]{0.5\linewidth}
\centering
\includegraphics[height=3cm,width=4cm]{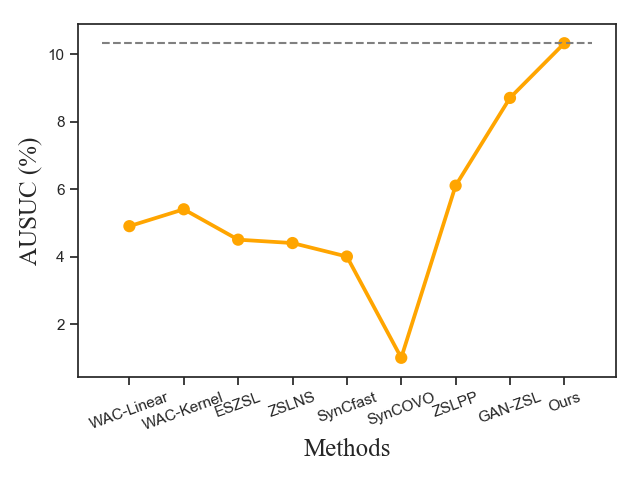}
\end{minipage}%
}%
\subfigure[NAB Dataset]{
\begin{minipage}[t]{0.5\linewidth}
\centering
\includegraphics[height=3cm,width=4cm]{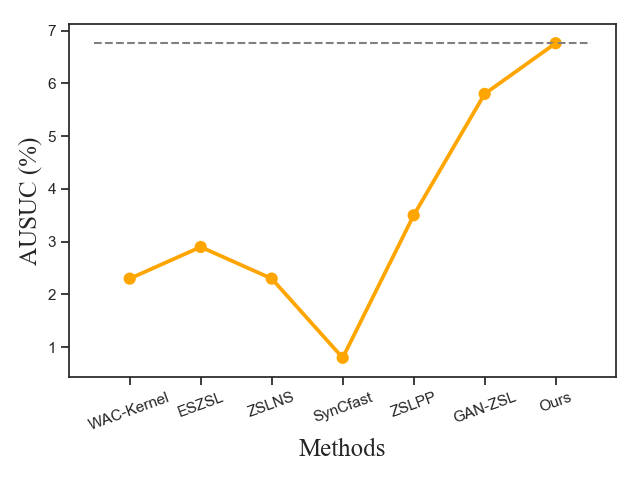}
\end{minipage}%
}%
\caption{AUSUC on two benchmarks datasets with SCE-split}
\label{SCE_AUSUC}
\end{figure}

A more general evaluation criterion is needed\cite{Zhu_2018_CVPR} since it is not enough to consider the performance of the unseen classes in ZSL. This metric takes into account the accuracy of both seen and unseen classes. A balance parameter is used to plot the curves of the seen classes and unseen classes (SUC). The area under SUC (AUSUC) is used to represent the generalization ability of the ZSL model. Fig \ref{SCS_AUSUC} and \ref{SCE_AUSUC} show the AUSUC scores between our method and other methods. The AUSUC score of our method is increased by 18.62\% and 16.55\%, respectively on two benchmarks datasets with SCE splitting compared with other highest methods. In SCS-split, 11.92\% improvement appears in the CUB dataset. On the NAB dataset, the AUSUC value slightly decreased, only 0.73\%.

\section{Conclusion}
In this paper, we propose a new method using knowledge sharing (KS) to augment class semantic to improve the accuracy of image classification in ZSL. Two major problems in ZSL task were solved: domain shift and semantic representation incomplete. Compared with the existing methods, the proposed method has achieved better results on CUB and NAB datasets. Our future work mainly includes two aspects. Firstly, semi-supervised learning is introduced to suppress domain drift further; Secondly, a knowledge map is added to better express semantic information.

\renewcommand\refname{Reference}
\bibliographystyle{plain}
\bibliography{ref}

\end{document}